\pdfoutput=1

\documentclass[11pt]{article}

\usepackage[]{acl}

\usepackage{times}
\usepackage{latexsym}

\usepackage[utf8]{inputenc} 
\usepackage[T1]{fontenc}    
\usepackage{hyperref}       
\usepackage{url}            
\usepackage{booktabs}       
\usepackage{amsfonts}       
\usepackage{nicefrac}       
\usepackage{microtype}      
\usepackage{xcolor}         
\usepackage{bm,bm,bbm,dsfont}
\usepackage{comment}
\usepackage{pgfplots}
\usepackage{arydshln}
\usepackage{pifont}
\usepackage{amsfonts}
\usepackage{multicol,multirow}
\usepackage{amsmath, amssymb, amsthm}
\usepackage[ruled]{algorithm2e}

\usepackage{bm,bbm,dsfont}
\pgfplotsset{width=7cm,compat=1.13}

\usepackage[T1]{fontenc}
\usepackage[utf8]{inputenc}

\usepackage{microtype}

\title{Fast Nearest Neighbor Machine Translation}

 \author{Yuxian Meng$^{\clubsuit}$,
  Xiaoya Li$^{\clubsuit}$, 
  Xiayu Zheng$^\blacktriangle$,
  Fei Wu$^{\spadesuit}$\\
  {\bf Xiaofei Sun$^{\spadesuit\clubsuit}$,
  Tianwei Zhang$^\blacklozenge$,
  Jiwei Li$^{\spadesuit\clubsuit}$
  }\\
    $^{\clubsuit}$Shannon.AI,
    $^{\spadesuit}$Zhejiang University \\
    $^\blacktriangle$Peking University, $^\blacklozenge$Nanyang Technological University\\
    \{yuxian\_meng, xiaoya\_li, xiaofei\_sun, jiwei\_li\}@shannonai.com\\
    xiayu\_zheng@pku.edu.cn, tianwei.zhang@ntu.edu.sg, wufei@zju.edu.cn
  }

\begin{document}
\maketitle

\begin{abstract}
Though nearest neighbor Machine Translation ($k$NN-MT) \citep{khandelwal2020nearest} has proved to introduce significant performance boosts over standard neural MT systems, it is prohibitively slow since it uses the entire reference corpus as the datastore for the nearest neighbor search. This means each step for each beam in the beam search has to search over the entire reference corpus. $k$NN-MT is thus two-orders slower than vanilla MT models,  making it hard to be applied to real-world applications, especially online services. In this work, we propose Fast $k$NN-MT to address this issue. Fast $k$NN-MT constructs a significantly smaller  datastore for the nearest neighbor search: for each word in a source sentence, Fast $k$NN-MT first selects its nearest token-level neighbors, which is limited to tokens that are the same as the query token. Then at each decoding step, in contrast to using the entire corpus as the datastore, the search space is limited to target tokens corresponding to the previously selected reference source tokens. This strategy avoids search through the whole datastore for nearest neighbors and drastically improves decoding efficiency. Without loss of performance, Fast $k$NN-MT is two-orders faster than $k$NN-MT, and is only two times slower than the standard NMT model. Fast $k$NN-MT enables the practical use of $k$NN-MT systems in real-world MT applications.
 \footnote{The code is available at \url{https://github.com/ShannonAI/fast-knn-nmt}}
\end{abstract}

\section{Introduction}
Machine translation (MT) is a fundamental task in natural language processing \citep{brown1993mathematics,och2003systematic}, and the prevalence of deep neural networks has spurred a diverse array of neural machine translation (NMT) models to improve translation quality \citep{sutskever2014sequence,bahdanau2014neural,vaswani2017transformer}.
 The recently proposed  
   $k$ nearest neighbor ($k$NN)  MT model \citep{khandelwal2020nearest} 
  has proved to 
  introduce significant performance boosts over standard neural MT systems. The basic idea behind $k$NN-MT is that at each decoding step, the model  
  is allowed to  
      refer to reference target tokens with similar translation contexts in a large datastore of cached examples.  The corresponding reference target tokens provide important insights on the translation token  likely to appear next. 
      
One notable limitation of $k$NN-MT is that it is prohibitively slow:
it uses the entire reference corpus as the datastore for the nearest neighbor search. 
This means 
each step for each beam in the beam search has to search over the entire reference corpus. $k$NN-MT is thus two-orders slower than vanilla MT models. 
The original paper of $k$NN-MT \citep{khandelwal2020nearest} suggests using fewer searching clusters, smaller beams and smaller datastores for generation speedup, but to achieve satisfactory results,  
carefully tuning on these factors under different tasks and datasets is still required 
 according to  analyses in \citep{khandelwal2020nearest}.  The computational overhead introduced by $k$NN-MT makes it hard to
 be
  deployed on real-world online services, which usually require both model performance and runtime efficiency.

In this work, we propose a fast version of $k$NN-MT -- Fast $k$NN-MT, to tackle the aforementioned issues. 
Fast $k$NN-MT constructs a significantly smaller  datastore for the nearest neighbor search:
for each word in a source sentence, Fast $k$NN-MT first selects its nearest token-level neighbors, which is limited to tokens of the same token type. 
Then at each decoding step, in contrast to 
consulting the 
entire corpus for nearest neighbor search, the datastore for
the currently decoding  token is limited within the tokens of reference targets corresponding to the previously selected reference source tokens, as shown in Figure \ref{fig:overview}. 
 The chain of mappings from the target token to the source token, then to its nearest source reference tokens, and last to corresponding  target reference tokens,  can be obtained using FastAlign \citep{dyer2013simple}. 
 
Fast $k$NN-MT provides several important advantages against vanilla $k$NN-MT in terms of speedup:
(1) the 
datastore
 in the KNN search is limited to target tokens corresponding to previously selected reference source tokens, instead of the entire corpus. This significantly improves decoding efficiency; 
(2) for source nearest neighbor retrieval, we propose to restrict the reference sources tokens that are the same as the query token, 
which further improves nearest-neighbor search efficiency. 
Without loss of performance, Fast $k$NN-MT is two-orders faster than $k$NN-MT, and is only two times slower than standard MT model. 
Under the settings of bilingual translation and domain adaptation, Fast $k$NN-MT achieves comparable results to $k$NN-MT, leading to a SacreBLEU score of 39.3 on WMT'19 De-En, 41.7 on WMT'14 En-Fr, and an average score of 41.4 on the domain adaptation task.


\section{Related Work}
\paragraph{Neural Machine Translation}~{}\\
Neural machine translation systems \citep{vaswani2017transformer,gehring2017convolutional,meng2019large} are often implemented by the sequence-to-sequence framework \citep{sutskever2014sequence} and enhanced with the attention mechanism \citep{bahdanau2014neural,luong-etal-2015-effective} which associates the current decoding token to the most semantically related part in the source side. 
At decoding time, beam search and its variants are used to find the optimal sequence \citep{sutskever2014sequence,li2016mutual}. 
The development of self-attention \citep{vaswani2017transformer} and pretraining \citep{devlin2018bert,lewis2019bart} has greatly motivated a line of works for more expressive MT systems.
These works include incorporating pretrained models \citep{zhu2020incorporating,guo2020incorporating}, designing lightweight model structures \citep{kasai2020deep,lioutas2020time,tay2020synthesizer,kasai2021finetuning,peng2021random}, handling multiple languages \citep{aharoni2019massively,arivazhagan2019massively,liu2020multilingual} and mitigating structural issues in Transformers \citep{wang2019learning,nguyen2019transformers,liu2020understanding,li2020sac,xiong2020layer} for more robust and efficient NMT systems.

\paragraph{Retrieval-Augmented Models}~{}\\
Retrieving and integrating auxiliary sentences has shown effectiveness in improving robustness and expressiveness for NMT systems.
\citep{zhang2018guiding} up-weighted the output tokens by collecting from the retrieved target sentences $n$-grams that align with the words in the source sentence, and \citep{bapna2019non} similarly retrieved $n$-grams but incorporated the information using gated attention \citep{cao2018encoding}.
\citep{tu2018learning} updated and stored the hidden representations of recent translation history in cache for access when new tokens are generated, so that the model can dynamically adapt to different contexts. 
\citep{gu2018search} leveraged an off-the-shelf search engine to retrieve a small subset of sentence pairs from the training set and then perform translation given the source sentence along with the retrieved pairs.
\citep{li2016one,farajian2017multi} proposed to retrieve similar sentences from the training set for the purpose of adapting the model to different input sentences.
\citep{bulte2019neural,jitao2020boosting} used fuzzy matches to retrieve similar sentence pairs from translation memories and augmented the source sentence with the retrieved pairs.
Our work is motivated by $k$NN-MT \citep{khandelwal2020nearest} and target improving the efficiency of $k$NN retrieval while achieving comparable translation performances.

Apart from machine translation, other NLP tasks have also benefited from retrieval-augmented models, such as language modeling \citep{khandelwal2019generalization}, question answering \citep{guu2020realm,lewis2020retrieval,lewis2020pre,xiong2020approximate} and dialog generation \citep{weston2018retrieve,fan2020augmenting,thulke2021efficient}.
Most of these works perform retrieval at the sentence level and treat the extracted sentences as additional input for model generation, whereas fast $k$NN-MT retrieves the most relevant tokens in the source side and fixes the probability distribution using the aligned target tokens at each decoding step.

\section{Background: $k$NN-MT}
Given an input sequence of tokens $\bm{x}=\{x_1,\cdots,x_n\}$ of length $n$, an MT model translates it into a target sentence in another language $\bm{y}=\{y_1,\cdots,y_m\}$ of length $m$.
A common practice to produce each token $y_i$ on the target side is to obtain a probability distribution over the vocabulary $p_\text{MT}(y_i|\bm{x},\bm{y}_{1:i-1})$ from the decoder and use beam search for generation. 
The combination of the complete source sentence and prefix of the target sentence $(\bm{x},\bm{y}_{1:i-1})$ is called {\it translation context}. 
$k$NN-MT interpolates this probability distribution with a multinomial distribution $p_\text{kNN}(y_i|\bm{x},\bm{y}_{1:i-1})$ derived from the $k$ nearest neighbors of the current translation context $(\bm{x},\bm{y}_{1:i-1})$ from a large scale datastore $\mathcal{S}$:
\begin{equation}
  \begin{aligned}
p(y_i|\bm{x},\bm{y}_{1:i-1})&=\lambda p_\text{kNN}(y_i|\bm{x},\bm{y}_{1:i-1})\\&+(1-\lambda)p_\text{MT}(y_i|\bm{x},\bm{y}_{1:i-1})
  \end{aligned}
  \label{knn-mt}
\end{equation}
More specifically, $k$NN-MT first constructs the datastore $\mathcal{S}$ using key-value pairs, where the key is the high-dimensional vector of the translation context produced by a trained MT model $f(\bm{x},\bm{y}_{1:i-1})$, and the value is the corresponding gold target token $y_i$, forming $\mathcal{S}=\{(\bm{k},v)\}=\{(f(\bm{x},\bm{y}_{1:i-1}),y_i)\}$. The context-target pairs may come from any parallel corpus.
Then, using the dense representation of the current translation context as query $\bm{q}=f(\bm{x}^\text{in},\bm{y}_{1:i-1})$ and $L_2$ distance as measure, $k$NN-MT searches through the entire datastore $\mathcal{S}$ to retrieve $k$ nearest translation contexts along with the corresponding target tokens $\mathcal{N}=\{\bm{k}_j,v_j\}_{j=1}^k$. Last, the retrieved set is transformed to a probability distribution by normalizing and aggregating the negative $L_2$ distances, $-d$, using the softmax operator with temperature $T$, which can be expressed as follows:
\begin{equation}
  \begin{aligned}
&  p_\text{kNN}(y_i|\bm{x}^\text{in},\bm{y}_{1:i-1})\\&=\frac{\sum_{(\bm{k}_j,v_j)\in\mathcal{N}}\mathds{1}_{y_i=v_j}\exp(-d(\bm{q},\bm{k}_j)/T)}{Z} \\
&  Z=\sum_{(\bm{k}_j,v_j)\in\mathcal{N}}\exp(-d(\bm{q},\bm{k}_j)/T)
  \end{aligned}
  \label{knn}
\end{equation}
Integrating Eq.(\ref{knn}) into Eq.(\ref{knn-mt}) gives the final probability of generating token $y_i$ for time step $i$.
Note that the above $k$NN search-interpolating process is applied to each decoding step of each beam, and each iteration needs to run on the full datastore $\mathcal{S}$. This gives a total time complexity of $\mathcal{O}(|\mathcal{S}|Bm)$, where $B$ is the beam size and $m$ is the target length.
In order for faster nearest neighbor search, $k$NN-MT leverages FAISS \citep{johnson2019billion}, an toolkit for efficient similarity search and clustering of dense vectors.

\section{Method: Fast $k$NN-MT}
\begin{figure*}[!t]
    \centering
    \includegraphics[width=1\textwidth]{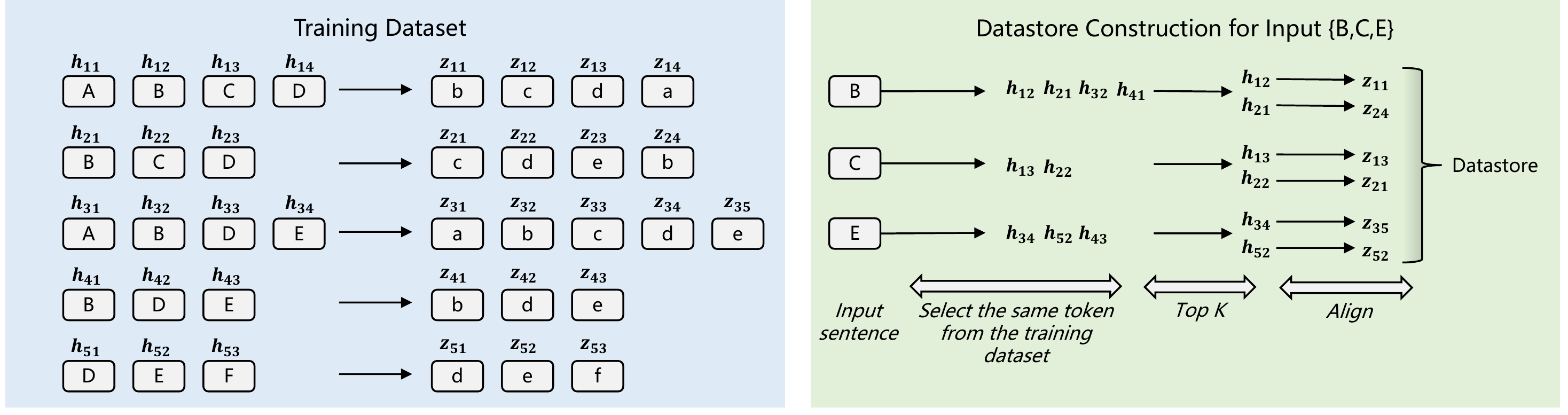}
    \caption{{\it Caching source and target tokens} (left, blue): Given a trained NMT model $f$ and the training corpus $\mathcal{D}_\text{train}$, we obtain representations for all source tokens $\bm{h}$ and target tokens $\bm{z}$ in the training set, which are the last layer outputs from $f$. 
    {\it Datastore construction} (right, green): Given a test example to translate, which is \{B,C,E\} in the example, we first navigate each source token to the tokens of the same type in the cache, e.g., $x_{12}$, $x_{21}$, $x_{32}$ and $x_{41}$ are identified for token B. Then, the top $c$ nearest neighbors for each source token are preserved according to the distance between the source token representation and candidate token representations, e.g., $x_{12}, x_{21}$ are selected for token B. 
    Last, the selected source tokens are aligned to their target tokens using FastAlign \citep{dyer2013simple}. For token B, the aligned target tokens are $y_{11}, y_{24}$. The collection of all aligned target tokens (along with their representations) constitutes the datastore for the current input \{B,C,E\}.}
    \label{fig:overview}
\end{figure*}

The time complexity of $k$NN-MT before search optimization is $\mathcal{O}(|\mathcal{S}|Bm)$\footnote{Time perplexity can slightly be alleviated when 
 approximate nearest neighbor search is used, with computation perplexity being not strictly $\mathcal{O}(|\mathcal{S}|)$. }, which is prohibitively slow when the size of the datastore $\mathcal{S}$ or the beam size $B$ is large.  
We propose strategies to address this issue.
The same as vanilla $k$NN-MT, fast $k$NN-MT system is built upon a separately trained MT encoder-decoder model.   
 To get a better illustration of how Fast $k$NN-MT works, we give a toy illustration in Figure \ref{fig:overview}. 
We use the capitalized characters to denote source tokens and lower-cased letters to denote target tokens. 
Given the training set, which is:
\begin{equation}
  \begin{aligned}
  (\bm{x}^{(1)},\bm{y}^{(1)}) &= (\{\text{A,B,C,D}\},\{\text{b,c,d,a}\})\\ (\bm{x}^{(2)},\bm{y}^{(2)}) &= (\{\text{B,C,D}\},\{\text{c,d,e,b}\})\\ (\bm{x}^{(3)},\bm{y}^{(3)}) &= (\{\text{A,B,D,E}\},\{\text{a,b,c,d,e}\})\\ (\bm{x}^{(4)},\bm{y}^{(4)}) &= (\{\text{B,D,E}\},\{\text{b,d,e}\})\\ (\bm{x}^{(5)},\bm{y}^{(5)}) &= (\{\text{D,E,F}\},\{\text{d,e,f}\})
  \end{aligned}
\end{equation}
in the toy example, 
an encoder-decoder model is  trained. 
Next, we wish to translate a source string $\{\text{B,C,E}\}$ at test time.

\subsection{Datastore Creation On the Source Side}
\label{sec:source}
Given a pretrained encoder-decoder model, and the training corpus, we first obtain representations for all source tokens and target tokens of the training set, which are the last layer outputs from the encoder-decoder model.
In the toy example,  representations  for source tokens \{A,B,C,D\}  in the first training example $(\{\text{A,B,C,D}\},\{\text{b,c,d,a}\})$ are respectively  $\bm{h}_{11}, \bm{h}_{12}, \bm{h}_{13}, \bm{h}_{14}$, and 
for target tokens \{b,c,d,a\} are respectively $\bm{z}_{11}, \bm{z}_{12}, \bm{z}_{13}, \bm{z}_{14}$. 
Given a test example to translate, which is \{B,C,E\} in the example,
we also obtain the representation for each of its constituent token, denoted by $\bm{h}_\text{B}, \bm{h}_\text{C}, \bm{h}_\text{E}$.
Next, we 
select nearest neighbor tokens for each source token, i.e., \{B, C, E\}. 
The nearest neighbor tokens are first limited to source tokens of the same token type as the query token.
For token B, tokens of the same token type are  
$x_{12}$,
$x_{21}$,  
$x_{32}$,
$x_{41}$. 
Similarly, for the token C in the test example, tokens of the same type are $\{x_{13}, x_{22}\}$;
for the token E, tokens of the same type are $\{x_{34}, x_{43}, x_{52}\}$.
One issue that stands out is that, for common words such as ``the'', there can be tens of millions of the same type tokens in the training corpus.
We thus need to further limit the number of nearest neighbors. 
Let $c$ denote the hyper-parameter that controls the number of nearest neighbors  for each token on the source side, which is set to 2 in the toy example. 
We rank all candidates based on the distance between the source token representation
(e.g., $\bm{h}_\text{B}, \bm{h}_\text{C}, \bm{h}_\text{E}$)
 and candidate token representations, 
and select the top $c$. Suppose that in the toy example, $x_{12}, x_{21}$ are selected for token B, 
$x_{13}, x_{23}$ are selected for token C,
 $x_{34}, x_{52}$ are selected for token E. 
The concatenation of selected candidates for all source tokens constitute the 
datastore on the source side, 
 which is $\mathcal{D}_\text{source} = \{x_{12}, x_{21}, x_{13}, x_{23}, x_{34}, x_{52}\}$ in the toy example. 
The  datastore creation for source tokens (e.g., \{B, C, E\}) can be run in parallel.

\subsection{Datastore Creation On the Target Side}
\label{sec:target}
For decoding, the model needs to refer to reference target tokens rather than source tokens.
We thus need to transform  
 $\mathcal{D}_\text{source}$ to a list of target tokens. 
 We use FastAlign \citep{dyer2013simple} toolkit to 
 achieve this goal.
 FastAlign maps source tokens to target tokens based on the IBM model \citep{och2003systematic}.
 Source tokens in   $\mathcal{D}_\text{source}$  that do not have correspondence on the target side 
 are abandoned. 
 Output target tokens from FastAlign form the datastore on the target side, denoted by  $\mathcal{D}_\text{target}$.
 In the toy example,   $x_{12}, x_{21}, x_{13}, x_{23}, x_{34}, x_{52}$ are respectively mapped to 
 $\mathcal{D}_\text{target} =\{ y_{11}, y_{24}, y_{12}, y_{21}, y_{35}, y_{52}\}$. 
The size of $\mathcal{D}_\text{target}$ is $c\times n$, where $n$ is the source length.

In practice, we first iterate over all examples in the training set, extracting all the source token representations and all the target token representations. Then, we build a separate token-specific cache $\mathcal{D}_v$ for each $v$ in vocabulary, which consists of (key, value) pairs where the key is the high-dimensional representation $\bm{h}$ and the value is a binary tuple containing the corresponding aligned target token along with its representation $\bm{z}$. 
Then we could map each source token of a given input sentence to its corresponding cache $\mathcal{D}_v$, and build the target-side datastore following the steps in Section \ref{sec:source} and Section \ref{sec:target}.
The process of caching source and target tokens is present in Algorithm \ref{algo:build_ds}.

\subsection{Decoding}
At the decoding time, the datastore for each decoding step is all limited to $\mathcal{D}_\text{target}$, within which $k$NN search is performed. 
Since tokens in $\mathcal{D}_\text{target}$ are not all related to the current decoding, 
 nearest neighbor search is performed to select the top $k$ candidates from $\mathcal{D}_\text{target}$ for each decoding step. 
For the nearest neighbor search here, we use the current representation $\bm{h}$ at the decoding time to query target representation 
$\bm{z}$ for target tokens in $\mathcal{D}_\text{target}$. 
The  selected nearest neighbors and their representations are used to compute the final word generation probability based on Eq.(\ref{knn-mt}) and Eq.(\ref{knn}). 

\begin{algorithm}[t]
    \footnotesize
      \SetKwInOut{Input}{Input}
      \SetKwInOut{Output}{Output}
      \Input{All sentence-pairs in training set: $(\bm{x}^{(1)}, \bm{y}^{(1)}), ..., (\bm{x}^{(N)}, \bm{y}^{(N)})$, vocabulary $\mathcal{V}$\\
             NMT encoder $f_e$, NMT decoder $f_d$, word alignment for each sentence-pair $(A^{(1)}, ..., A^{(N)})$ \\
             Input for test: $\bm{x}$}
      \Output{Target datastore $\mathcal{D}_\text{target}(\bm{x})$ for  $\bm{x}$ }
            $\mathcal{D}_\text{target}(\bm{x}) \leftarrow \emptyset$ \qquad\qquad\qquad\quad $\blacktriangleright$ initialize  the datastore $\mathcal{D}_\text{source}(\bm{x}) $ for $\bm{x}$ \\

      \For{$v \leftarrow 1$ \KwTo $\mathcal{V}$}{ 
          $\mathcal{D}_v \leftarrow \emptyset$ \qquad\qquad\qquad\qquad $\blacktriangleright$ initialize the (key, value) datastore for each word in the vocabulary
      }
      \% {\it Caching Source and Target Tokens}: \\
      \For{$i \leftarrow 1$ \KwTo $N$}{
          $n_i \leftarrow$ length of $\bm{x}^{(i)}$ \\
          $m_i \leftarrow$ length of $\bm{y}^{(i)}$ \\
          $[\bm{h}_1, ..., \bm{h}_{n_i}] \leftarrow f_e(\bm{x}^{(i)})$ \qquad\qquad\qquad\qquad $\blacktriangleright$ computing representations for each source word \\
          $[\bm{z}_1, ..., \bm{z}_{m_i}] \leftarrow f_d(\bm{x}^{(i)}, \bm{y}^{(i)})$ \qquad\qquad\qquad $\blacktriangleright$ computing representations for each target word\\
          \For{$j \leftarrow 0$ \KwTo $n_i$}{
              \For{$k \leftarrow 0$ \KwTo $m_i$}{
                    \If{$(j, k) \in A^{(i)}$}{
                         add $(\bm{h}_i, (\bm{z}_k, y^{(i)}_k))$ to $\mathcal{D}_{x^{(i)}_{j}}$ \\
                         \% {\it key of $\mathcal{D}_{x^{(i)}_{j}}$ are all hidden representations $\bm{h}_i$ for tokens of the same token type as  $x^{(i)}_{j}$} 
                   }}
          }
      }
      \% {\it Generate Datastore for the test input $\bm{x}$}: \\
      $n_x \leftarrow$ length of $\bm{x}$  \\
      \For{$k \leftarrow 1$ \KwTo $N$}{
      		obtain hidden representation for the $k$-th token $\bm{h}_k$ \\
		$\{\bm{z}, y\}$ $\leftarrow$ top $c$ items from $\mathcal{D}_{x_k}$ using $\bm{h}_k$ as the query\\
      		add $\{\bm{z}, y\}$ to $\mathcal{D}_\text{target}(x)$ 
      }
      \caption{Constructing Datastore for a Test Input $\bm{x}$.}
      \label{algo:build_ds}
\end{algorithm}

\subsection{Quantization}\label{quantize}
Although the prohibitive computational cost issue of $k$NN-MT has been addressed, the intensive memory for datastore remains a problem, as we wish to cache 
all source and target representations of the entire training set. 
Additionally, 
frequently accessing Terabytes of data is also extremely time-intensive. 
To address this issue, we propose to use product quantization (PQ) \citep{jegou2010product} to compress the high-dimensional representations of each token. Formally, given a vector $\bm{x} \in \mathbb{R}^{D}$, we represent it as the concatenation of $M$ subvectors: $\bm{x}=[\bm{x}^1, \bm{x}^2, ..., \bm{x}^M]$, where all subvectors have the same dimension $d=D/M$. The product quantizer $q$ consists of $M$ sub-quantizers $q_1, ..., q_M$, each maps a subvecor $\bm{x}^m \in \mathbb{R}^d$ to a codeword $\bm{c}$ in a subspace codebook with $n$ codewords: $\mathcal{C}^m=\{\bm{c}^m_1, \bm{c}^m_2, ..., \bm{c}^m_n\}$. The objective function of PQ is:
\begin{equation}
    \min_{q^1, ..., q^M} \sum_{\bm{x}} \sum_{m=1}^M \|{\bm{x}^m-q_m(\bm{x}^m)}\|^2
\end{equation}
Therefore each $\bm{x}$ is mapped to its nearest codeword in the Cartesian product space $\mathcal{C}=\mathcal{C}^1 \times ... \times \mathcal{C}^M $. If each subspace codebook $\mathcal{C}^m$ has $n$ codewords, then the Cartetian product space $\mathcal{C}$ could represent $n^m$ $D$-dimensional codewords with only $n \times m$ $d$-dimensional vectors, thus significantly eased the memory issue.
With the quantization technique, 
we are able to compress 
 each token representation  to 128-bytes. For example, for the WMT19 En-De dataset, the memory size is reduced from 3.5TB to  108GB.

\subsection{$k$NN Retrieval Details}
In practice, executing exact nearest neighbor search over millions or even billions of tokens could be time-consuming. 
Hence we use FAISS \citep{johnson2019billion} for fast approximate nearest neighbor search. All token representations are quantized to 128-bytes. Recall that we build a token-specific datastore $\mathcal{D}_v$ for each $v$ in vocabulary. We do brute force search for tokens whose frequency $n_v$ is lower than 30000. For those tokens whose frequency is larger than 30000, the keys are stored in clusters to speed up search. The number of clusters for token $v$ is set to $\min(4 \times \sqrt{n_v}, n_v/30)$. To learn the cluster centroids, we use at most 5M keys for each token $v$. During inference, we query the datastore for $k=512$ neighbors through searching 32 nearest clusters.

\subsection{Discussions on Comparisons to Vanilla $k$NN-MT}
The speedup of Fast $k$NN-MT lies in the following three aspects: \\
{\bf (1)} For nearest neighbor retrieval on the source side, we first restrict the reference  tokens that are the same as the query token.
This strategy significantly narrows down the search space to roughly $|\mathcal{S}|/\text{mid(F)}$ times, where $|\mathcal{S}|$ denotes the number of tokens in the corpus, and $\text{mid}(F)$
denotes the medium word frequency in the corpus. \\
{\bf (2)} The nearest neighbor search for all 
source tokens 
on the source side can be run in parallel, which is also a key speedup over  $k$NN-MT. 
For vanilla $k$NN-MT, 
 $k$NN search is performed on the target side and has to be auto-regressive: 
 the representation for the current decoding step, which is used for the $k$NN search over the entire corpus, 
  relies on previously generated tokens.
Therefore, the $k$NN search for the current step has to wait for the finish of $k$NN searches for all previous generation steps.\\
{\bf (3)} 
On the target side,
the 
datastore
 in the $k$NN search is limited to target representations corresponding to  selected reference source tokens. 
 Though the nearest neighbor search in the decoding process is auto-regressive and thus cannot be run in parallel, the running cost is fairly low: 
 recall that the size of $\mathcal{D}_\text{target}$ is $c\times n$. Across all settings, the largest value of $c$ is set to 512.
 The size of $\mathcal{D}_\text{target}$ is roughly 15k. Performing nearest neighbor searches among 15k candidates is relatively cheap for NMT, and is actually cheaper than the softmax operation for word prediction, where the vocabulary size is usually around 50k. \\
The combination of all these aspects leads to Fast $k$NN-MT two orders of magnitude faster than vanilla $k$NN-MT.

\section{Experiments}

\subsection{Bilingual Machine Translation}\label{wmt-section}
We conduct experiments on two bilingual machine translation datasets: WMT'14 English-French and WMT'19 German-English. To create the datastore, we follow \citep{ng2019facebook} to apply language identification filtering, keeping only sentence pairs with correct languages on both sides. We also remove sentences longer than 250 tokens and sentence pairs with a source/target ratio exceeding 1.5. For all datasets, we use the standard Transformer-base model provided by FairSeq \citep{ott2019fairseq} library.\footnote{\url{https://github.com/pytorch/fairseq/tree/master/examples/translation}} The model has 6 encoder layers and 6 decoder layers. The dimensionality of word representations is 1024, the number of multi-attention heads is 16, and the inner dimensionality of feedforward layers is 8192. 
Particularly, following \citep{khandelwal2020nearest}, the model for WMT'19 German-English has also been trained on over 10 billion tokens of extra backtranslation data as well as fine-tuned on newstest test sets from previous years.
We report the SacreBLEU scores \citep{post-2018-call} for comparison.\footnote{\url{https://github.com/mjpost/sacrebleu}}
Table \ref{wmt-result} shows our results on the two NMT datasets. The proposed Fast $k$NN-MT model is able to achieve slightly better results to the vanilla $k$NN-MT model on WMT'19 German-English, and competitive results on WMT'14 English-French, with less $k$NN search cost.

\begin{table}[t]
\centering
    \begin{tabular}{lll}
    \toprule
     Model & De-En & En-Fr  \\\midrule
    base MT              & 37.6               & 41.1                 \\
    +$k$NN-MT            & 39.1$_{(+1.5)}$    & 41.8$_{(+0.7)}$      \\
    +fast $k$NN-MT       & 39.3$_{(+1.7)}$    & 41.7$_{(+0.6)}$      \\       
    \bottomrule
    \end{tabular}
\caption{SacreBLEU scores on WMT'19 De-En and WMT'14 En-Fr.}
\label{wmt-result}
\end{table}

\subsection{Domain Adaptation}
\label{experiment-domain-adapt}
We also measure the effectiveness of the proposed Fast $k$NN-MT model on the domain adaptation task. We use the multi-domain datasets which are originally provided in \citep{koehn2017six} and further cleaned by \citep{aharoni2020unsupervised}. These datasets include German-English parallel data
for train/validation/test sets in five domains: Medical, Law, IT, Koran and Subtitles. We use the trained German-English model introduced in Section \ref{wmt-section} as our base model, and further build domain-specific datastores to evaluate the performance of Fast $k$NN-MT on each domain. Table \ref{domain-adapt-result} shows that Fast $k$NN-MT achieves comparable results to vanilla $k$NN-MT on domains of Medical (54.6 vs. 54.4), IT (45.9 vs. 45.8) and Subtitles (31.9 vs. 31.7), and outperforms vanilla $k$NN-MT on the domain of Koran (21.2 vs. 19.4). The average score of Fast $k$NN-MT (41.7) is on par with the result of \citep{aharoni2020unsupervised} (41.3), which trains domain-specific models and reports in-domain results.

Following \citep{khandelwal2020nearest}, we also carry out experiments under the out-of-domain and multi-domain settings and report the results on Table \ref{domain-adapt-result}. ``+ WMT19' datastore'' shows the results for retrieving neighbors from 770M tokens of WMT’19 data that the model has been trained on, and ``+ all-domain datastore'' shows the results where the model is trained on the multi-domain datastore from all six settings. The BLEU improvement is much smaller on the out-of-domain setup compared to the in-domain setup, illustrating that the proposed framework relies on in-domain data to retrieve valuable contexts. For the multi-domain setup, the performance for all six domains generally remains the same and only a small drop of the average score is witnessed. This shows that the Fast $k$NN-MT framework is robust to a massive amount of out-of-domain data and is able to retrieve the context-related information from in-domain data.

\begin{table*}[!t]
    \centering
    \scalebox{0.9}{
    \begin{tabular}{lllllll}
    \toprule
    Model & Medical & Law & IT & Koran & Subtitles & Avg.  \\\midrule
    \citep{aharoni2020unsupervised} &  54.8 & 58.8 & 43.5 & 21.8 & 27.4 & 41.3  \\
    base MT  &   39.9  &  45.7  &  38.0  &  16.3  & 29.2  & 33.8                    \\
    +in-domain datastore: \\
    ~~~~$k$NN-MT  &  54.4$_{(+14.5)}$ & 61.8$_{(+16.1)}$ & 45.8$_{(+7.8)}$ & 19.4$_{(+3.1)}$  & 31.7$_{(+2.5)}$ & 42.6$_{(+8.8)}$        \\
    ~~~~fast $k$NN-MT  & 53.6$_{(+13.7)}$ &  56.0$_{(+10.3)}$  & 45.5$_{(+7.5)}$ & 21.2$_{(+4.9)}$  &  30.5$_{(+1.3)}$  &  41.4$_{(+7.6)}$\\ 
    \cmidrule(r){1-7}
    +WMT19' datastore \\
    ~~~~$k$NN-MT  &  40.2$_{(+0.3)}$ & 46.7$_{(+1.0)}$ & 40.3$_{(+2.3)}$ & 18.0$_{(+1.7)}$  & 29.2$_{(+0.0)}$ & 34.9$_{(+1.1)}$        \\
    ~~~~fast $k$NN-MT  & 41.5$_{(+1.6)}$ &  45.9$_{(+0.2)}$  & 41.0$_{(+3.0)}$ & 17.6$_{(+1.3)}$  &  29.2$_{(+0.0)}$  &  35.0$_{(+1.2)}$\\   
    +all-domains datastore \\
    ~~~~$k$NN-MT  &  54.5$_{(+14.6)}$ & 61.1$_{(+15.4)}$ & 48.6$_{(+10.6)}$ & 19.2$_{(+2.9)}$  & 31.7$_{(+2.5)}$ & 43.0$_{(+9.2)}$        \\
    ~~~~fast $k$NN-MT  & 53.2$_{(+13.3)}$ &  54.6$_{(+8.9)}$  & 46.4$_{(+8.4)}$ & 19.5$_{(+3.2)}$  &  29.7$_{(+0.5)}$  &  40.7$_{(+6.9)}$\\      
    \bottomrule
    \end{tabular}
    }
\caption{SacreBLEU results for domain adaptation.}
\label{domain-adapt-result}
\end{table*}

\begin{table*}[!t]
\centering
\scalebox{0.9}{
    \begin{tabular}{p{4cm}p{5.5cm}p{5.5cm}}
    \toprule
     & {\bf Source} & {\bf Target}  \\
    \hline
    original sentence pair & Zwei Fisch@@ betriebe haben ihre Tätigkeit eingestellt . & Two fish establi@@ sh@@ ments have ce@@ ased their activities .  \\
    \hline
    reference of "Zwei"  &   {\color{red}Zwei} Gemein@@ schaft@@ sher@@ steller in Griechenland , die bei der vor@@ ausge@@ gangenen Untersuchung mit@@ gearbeitet hatten , gaben ihre Tätigkeit auf . 
             &   Fur@@ ther@@ more , {\color{red}two} Community producers in Greece who took part in the previous investigation ce@@ ased their activity .                \\
    \hline 
    reference of "Fisch@@"  &  - fünf auf die {\color{red}Fisch@@} industrie , 
              & - five to representatives of the {\color{red}fi@@ sh@@ ery} industries ,                        \\
    \hline
    reference of "betriebe"  &  ( 4 ) Drei Fleisch@@  {\color{red}betriebe} mit Übergangs@@ regelung haben erhebliche Anstrengungen unter@@ nommen , neue Betriebs@@ anlagen zu bauen . 
              &  ( 4 ) Three meat establi@@ sh@@ ments on the list of {\color{red}establi@@ sh@@ ments} in transition have made considerable efforts to build new facilities .                     \\
    \hline 
    reference of "haben", "ihre", "Tätigkeit", "eingestellt" and "."  & Einige Betriebe {\color{red}haben ihre Tätigkeit eingestellt .}
              & Cer@@ tain establi@@ sh@@ ments {\color{red}have ce@@ ased their activities .}                     \\
    \bottomrule
    \end{tabular}
    }
\caption{A  test sentence pair from the Law domain. We show the original sentence pair for test (the first row), the nearest-neighbor tokens 
on the source side
 along with the sentences that retrieved tokens reside in (the second column), and the aligned target tokens 
 extracted from FastAlign, 
  along with  sentences in which  target tokens reside in (the third column). The retrieved tokens are in red.}
\label{examples}
\end{table*}

\subsection{Analysis}
\begin{figure*}[t]
    \centering
    \includegraphics[width=1\textwidth]{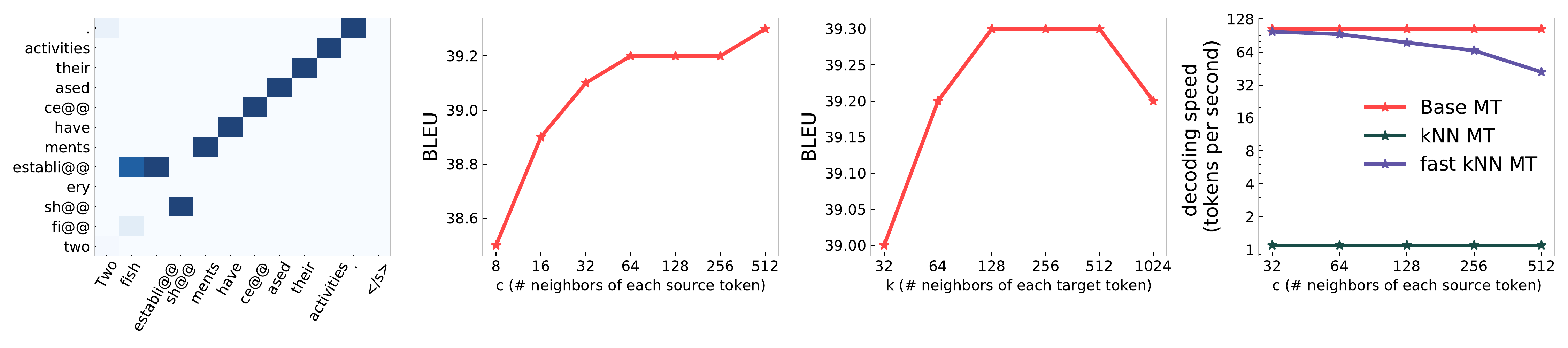}
    \caption{{\it First figure}: the similarity heatmap between the gold target tokens and the retrieved target neighbors. 
    {\it Second figure}: BLEU scores with respect to different $c$, the number of nearest neighbors on the source side for each source token.
     {\it Third figure}: BLEU scores with respect to different $k$, the number of nearest neighbors on the target side for each target token.
      {\it Forth figure}: Speed comparison between base MT, $k$NN-MT and fast $k$NN-MT.}
    \label{fig:merge}
\end{figure*}

\paragraph{Examples}
To visualize the effectiveness of the proposed Fast $k$NN-MT model, we randomly choose an example from the test set of the Law domain. 
Table \ref{examples} shows the test sentence, the retrieved nearest neighbor tokens on the source side, and the corresponding target tokens.  The first figure in Figure \ref{fig:merge} demonstrates the similarity heatmap between the gold target tokens and the selected target neighbors. We can see that the retrieved target nearest tokens are highly correlated with the ground-truth target tokens, exhibiting the ability of Fast $k$NN-MT to accurately extract nearest reference tokens at each decoding step.

\paragraph{The Effect of the Number of neighbors per token on the source side}
We queried the datastore for nearest $c$ neighbors for each source token. Intuitively, the larger the $c$ is, the more likely
the model could recall the nearest neighbors on the target side. The second figure in Figure\ref{fig:merge} verifies this point: the model performance increases drastically when $c$ increases from 8 to 64, and then continues increasing as $c$ is up to 512.

\paragraph{The Effect of the Number of neighbors per token on the target side}
Fast $k$NN-MT selects top $k$ nearest neighbors at each decoding step for computing the probability $p_\text{kNN}$ in Eq.(\ref{knn}).
The third figure in Figure \ref{fig:merge} shows that the model performance first increases and then decreases when we continue enlarging the value of $k$, with $c$ fixed at 512, which is consistent with the observation in \citep{khandelwal2020nearest}. This is because that using neighbors that are too far away from the ground-truth target token adds noise to the model prediction, and thus hurts the performance.

\paragraph{Speed comparison}
When the beam size is fixed, the time complexity of Fast $k$NN-MT is mainly 
controlled by the number of retrieved neighbors $c$ for each source token.\footnote{$k$ plays a minor role to the overall time complexity because each search on the target side is performed within a total amount of $cn$ tokens, which is negligible compared to the time cost spent on the source side.} The last figure in Figure \ref{fig:merge} shows the speed comparison between base MT, $k$NN-MT and fast $k$NN-MT when we vary the value of $c$. 
Fast $k$NN-MT decoded nearly as fast as the vanilla MT model when $c$ is small.  When $c$ reaches 512, 
 $k$NN-MT is about two times slower than the vanilla MT model.
By contrast, vanilla $k$NN-MT is two order of magnitude slower than base MT and Fast $k$NN-MT regarding the decoding speed. This is because Fast $k$NN-MT substantially restricts the search space during decoding, whereas vanilla $k$NN-MT has to execute $k$NN search over the entire datastore at each decoding step.

\paragraph{Similarity function}
We have tried different similarity functions when retrieving $c$ nearest neighbors on source side and computing the $k$NN distribution. These functions include cosine similarity, inner product and $L_2$ distance, the SacreBLEU scores for which are respectively 39.2, 39.1 and 38.8 on WMT'19 German-English,
showing that cosine similarity is a better measure for representation distance than $L_2$ distance and inner product. 


\paragraph{Effect of quantization}
\begin{table}[t]
\centering
\scalebox{0.7}{
    \begin{tabular}{lllllll}
    \toprule
    Model & Medical & Law & IT & Koran & Subtitles & Avg.  \\\midrule
    fast $k$NN-MT  & 53.6 &  61.0  &   45.5  & 21.2  &  31.9  &  41.7               \\     
    ~~+full-precision &  53.8 & 61.1 & 45.8 & 21.3  & 30.7 & 41.5           \\
    \bottomrule
    \end{tabular}
    }
\caption{SacreBLEU scores on domain adaptation when using full precision.}
\label{table:quantization}
\end{table}
Due to the memory issue, we applied quantization to compress the high-dimensional representation of each token in the training set.
We investigate how quantization would affect model performances. 
As shown in Table \ref{table:quantization}, quantization has minor side effects in terms of BLEU scores, 
and when we use full precision instead of quantization, the average BLEU score only increases 0.1, which suggests that computing similarity using compressed vectors is a viable trade-off between memory usage and model performance.

\section{Conclusion}
In this work, we propose a fast version of $k$NN-MT -- Fast $k$NN-MT -- to address the runtime complexity issue of the vanilla $k$NN-MT.
During decoding, Fast $k$NN-MT constructs a significantly smaller  datastore for the nearest neighbor search:
for each word in a source sentence, Fast $k$NN-MT selects its nearest tokens from a large-scale cache. The selected tokens are the same as the query token. 
 Then at each decoding step, in contrast to 
using the entire datastore, the 
search space is limited to 
  target tokens corresponding to the previously selected reference source tokens. 
Experiments demonstrate that this strategy drastically improves decoding efficiency while maintaining model performances compared to vanilla $k$NN-MT under different settings including bilingual machine translation and domain adaptation. Comprehensive ablation studies are performed to understand the behavior of each component in Fast $k$NN-MT. 
In future work, we plan to further improve the efficiency of Fast $k$NN-MT by applying clustering techniques to build the datastore.

\section*{Acknowledgement}
This work is supported by the Science and Technology Innovation 2030 - “New Generation Artificial Intelligence” Major Project (No. 2021ZD0110201) and the Key R \& D Projects of the Ministry of Science and Technology (2020YFC0832500).
We would like to thank anonymous reviewers for their comments and suggestions.   

\bibliography{custom}
\bibliographystyle{acl_natbib}



\end{document}